          [2001/04/25 1.1 (PWD)]
\documentclass[a4paper,twocolumn]{esapub2005} 
\usepackage[utf8]{inputenc}
\usepackage{times}
\usepackage[square,numbers]{natbib}
\usepackage{url}
\usepackage{xcolor}
\usepackage{amsmath}
\usepackage{amssymb}
\usepackage{amsthm}
\usepackage{booktabs}
\usepackage{layouts}
\usepackage{siunitx}
\usepackage{multirow}
\usepackage{graphicx}
\usepackage{placeins}
\usepackage{optidef}
\usepackage{hyperref}
\usepackage{stmaryrd}

\title{Quadrupeds for Planetary Exploration: Field Testing Control Algorithms on an Active Volcano}
\author[1]{Shubham Vyas}
\author[1]{Franek Stark}
\author[1]{Rohit Kumar}
\author[2]{Hannah Isermann}
\affil[1]{Robotics Innovation Center, DFKI GmbH, Bremen, Germany, \small{\{shubham.vyas, franek.stark, r.kumar, dennis.mronga\}@dfki.de}}
\author[3]{Jonas Haack}
\author[3]{Mihaela Popescu}
\author[1,3]{Jakob Middelberg}
\affil[2]{German Aerospace Center, Institute of Space Systems, Bremen, Germany, hannah.isermann@dlr.de}
\affil[3]{University of Bremen, Germany, \small{\{jhaack, mpopescu, middelberg\}@uni-bremen.de}}
\author[1]{Dennis Mronga}
\author[1,3]{Frank Kirchner}

\begin{document}


\maketitle

\begin{abstract}
Missions such as the Ingenuity helicopter have shown the advantages of using novel locomotion modes to increase the scientific return of planetary exploration missions. Legged robots can further expand the reach and capability of future planetary missions by traversing more difficult terrain than wheeled rovers, such as jumping over cracks on the ground or traversing rugged terrain with boulders. To develop and test algorithms for using quadruped robots, the AAPLE project was carried out at DFKI. As part of the project, we conducted a series of field experiments on the Volcano on the Aeolian island of Vulcano, an active stratovolcano near Sicily, Italy. The experiments focused on validating newly developed state-of-the-art adaptive optimal control algorithms for quadrupedal locomotion in a high-fidelity analog environment for Lunar and Martian surfaces. This paper presents the technical approach, test plan, software architecture, field deployment strategy, and evaluation results from the Vulcano campaign.
\end{abstract}

\section{Introduction}
The recent success of the Ingenuity helicopter on Mars \cite{balaram2021ingenuity} has highlighted the transformative potential of novel locomotion modes in planetary exploration. By augmenting traditional rover-based missions with aerial mobility, Ingenuity has enabled access to previously unreachable areas, significantly increasing scientific return. Inspired by this paradigm shift, legged robots—specifically quadrupeds—present an opportunity to further expand the reach and capability of future planetary missions. Unlike wheeled rovers, which are limited by terrain constraints such as loose regolith, steep slopes, and large boulders, quadrupeds offer dynamic mobility that can accommodate irregular, non-flat surfaces, allowing them to jump across fissures, climb over rocks, and adapt to unstable ground conditions. These capabilities make them promising candidates for exploration of the Moon, Mars, and other planetary bodies with similarly challenging terrain.

\begin{figure}[t]
    \centering
    \includegraphics[width=\linewidth]{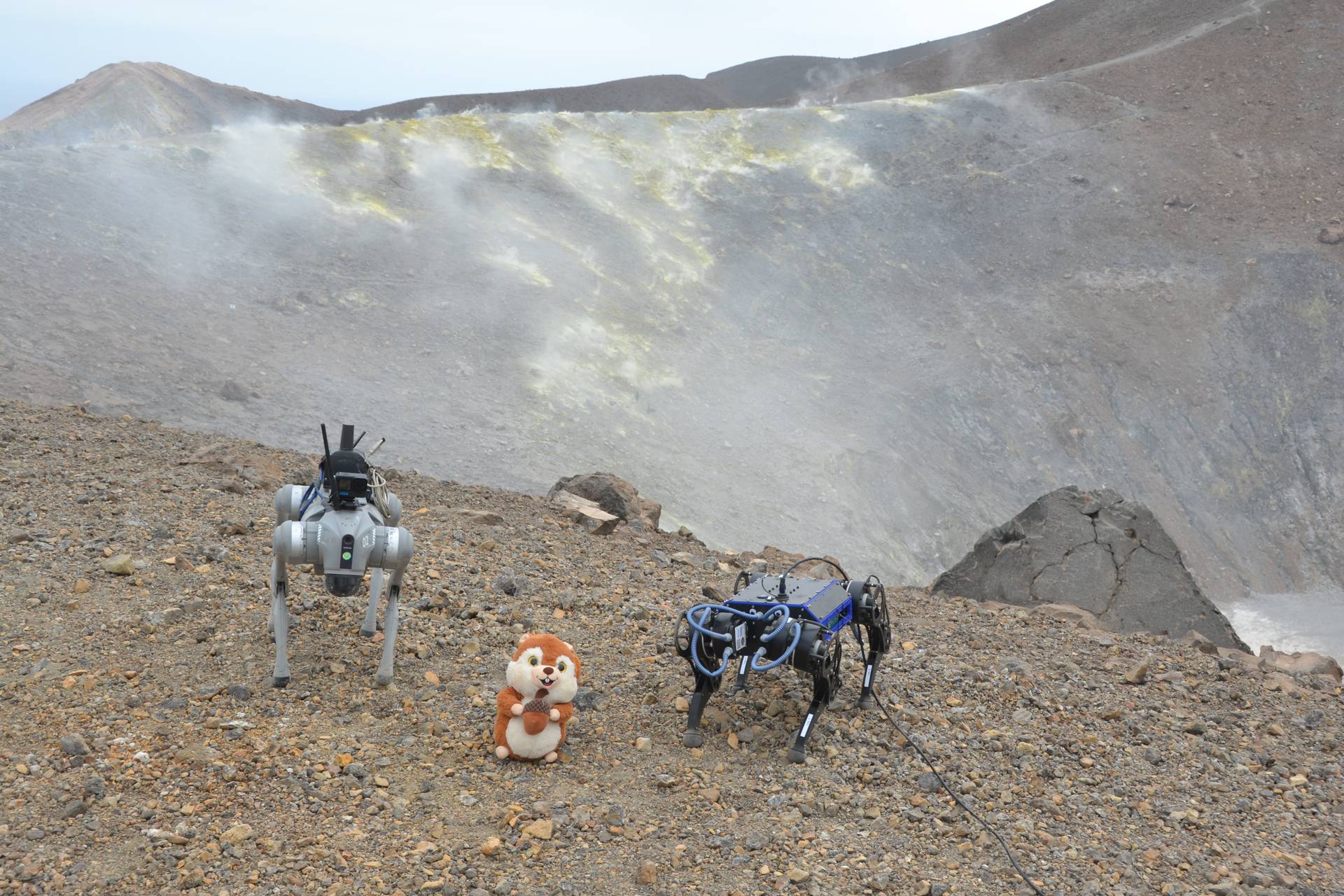}
    \caption{Two quadrupedal robots used during the field tests on Vulcano: Unitree Go2 (left) and Custom built B12 (right).}
    \label{fig:robots_on_vulcano}
\end{figure}

The AAPLE project at DFKI~\cite{dfkibremenaaple} was carried out to explore the limits of legged locomotion for planetary exploration. As part of the project, to evaluate the readiness and performance of quadrupedal locomotion control strategies under realistic extraterrestrial conditions, we conducted a series of field experiments on the Volcano on the Aeolian island of Vulcano, an active stratovolcano near Sicily, Italy. These field tests were performed in the context of the PETRAS Vulcano Summer School 2024 on Planetology, Exploration, Terrestrial Analogs, Robotics, Astrobiology, and Spectroscopy. The summer school was a collaborative effort between University of Perugia (UniPG), TU Bergakademie Freiberg (TUBAF), IU International University of Applied Sciences (IUBH), European Astrobiology Institute (EAI), German Research Center for Artificial Intelligence (DFKI), and various sections of the German Aerospace Center (DLR). The field tests at the summer school included 9 members from DFKI, organized into two robot groups: 6 for quadrupedal robots and 3 for the hybrid wheel-leg rover Coyote~\cite{sonsalla2015coyote}.

Vulcano serves as a high-fidelity analog environment for lunar and Martian surfaces, offering terrain composed of loose, granular regolith, frequent elevation changes, and geological features analogous to those expected in future exploration missions. These field trials provided an essential validation step to increase the technology readiness level of the developed GNC sub-system for legged locomotion. The experiments focused on validating newly developed state-of-the-art adaptive optimal control algorithms for quadrupedal locomotion. These algorithms employ model predictive control (MPC) and whole-body optimization techniques that allow the robot to dynamically adapt to uncertain and highly variable terrain in real time. During the campaign, the robot executed a variety of locomotion tasks, including remotely operated traversal over uneven volcanic ash, inclined rocky slopes, and negotiating discontinuous terrain with crevices and obstacles.

This paper presents the technical approach, test plan, software architecture, field deployment strategy, and evaluation results from the Vulcano campaign. By validating control algorithms in a geologically and logistically relevant setting, we advance the readiness of quadrupedal robots for planetary exploration. Our results support the integration of legged mobility into future lunar and Martian mission architectures, where it can serve as a valuable complement to rovers and aerial systems in accessing extreme terrain and maximizing scientific return. Lessons learned from this field test will inform future field campaigns and space mission payload designs, where robustness and ease of deployment are essential.

\section{Logistics}
\label{sec:logistics}

Logistical planning was critical to the success of the field campaign on Vulcano. This section outlines the key logistical considerations addressed during the preparation and execution phases of the field tests. Transportation of the Unitree Go2 quadruped robot presented significant challenges due to regulatory constraints around shipping large lithium batteries. We collaborated with \textit{mybotshop.de}~\cite{mybotshop}, the original equipment supplier, who provided specialized expertise in handling and transporting research robots across international borders, ensuring compliance with shipping regulations while maintaining equipment integrity.

\subsection{Travel and Permits}
For the travel of the researchers and robots, separate arrangements were made. The robots were shipped earlier so that they could arrive a few days before the team at Sicily and arrive on the Vulcano island when a part of the team was already there. A part of the team traveled two days earlier to Vulcano to prepare for the reception of the robots as the shipping company couldn't provide exact arrival time, but rather a time window of a few days due to ferry logistics. The team traveled from Germany to Sicily and then via ferry to Vulcano. This approach minimized time risks and ensured that the robots were available for deployment upon the team's arrival. All necessary permits for conducting research on Vulcano were arranged by the organizers of the PETRAS Vulcano Summer School, providing a unique opportunity for robotics researchers who typically do not have experience in securing such permits, unlike the experienced geologists. This support ensured compliance with regulations governing activities in protected areas and facilitated smooth execution of the field campaign.

On the site, the researchers and scientists from the PETRAS summer school taught us about the volcanic features and the geological history of the island. The team also had to be cautious about the active volcanic nature of the site, which required constant monitoring of volcanic activity reports and adherence to safety protocols established by scientists. A key factor to the success of the tests was the information and support provided by the local scientists and volcanologists, who were part of the summer school as this knowledge is not common for roboticists.

\subsection{Software and Network Management}

\begin{figure}[htpb]
    \centering
    \includegraphics[width=\linewidth]{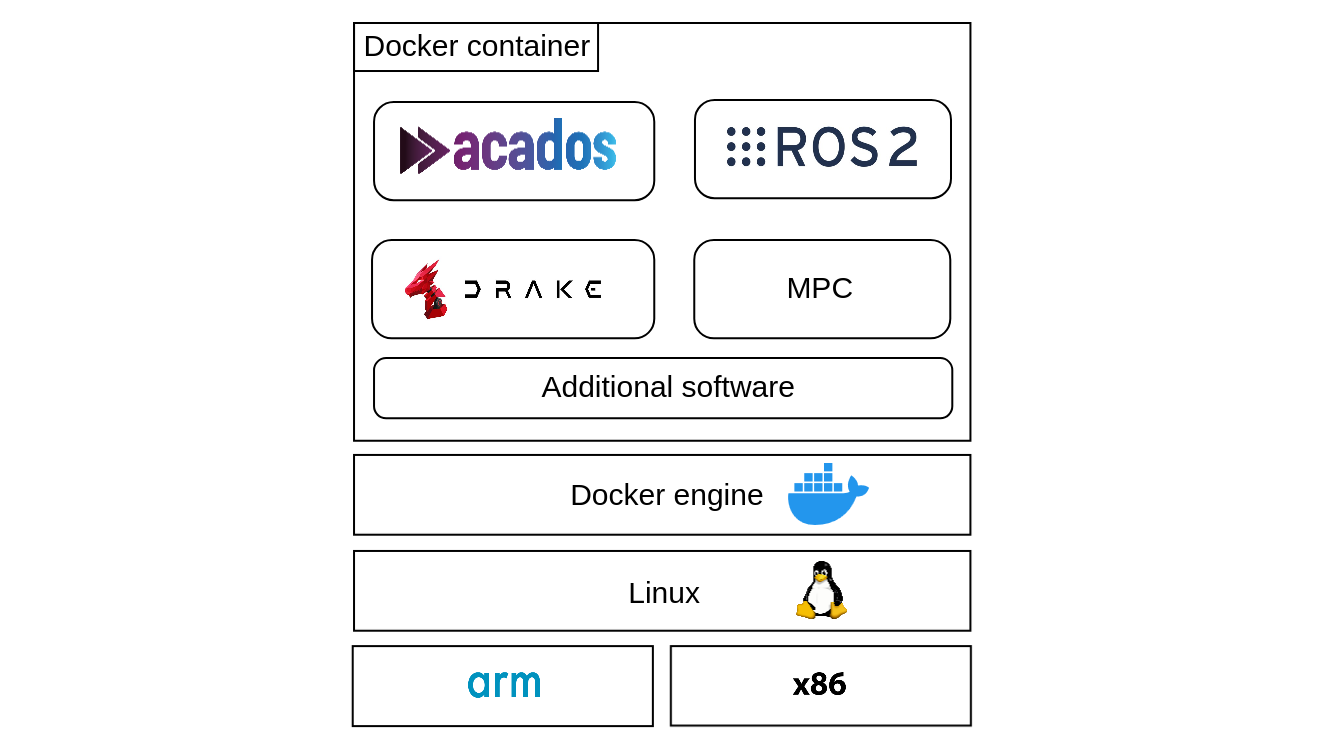}
    \caption{Software Stack used during the field tests on Vulcano.}
    \label{fig:sw_stack}
\end{figure}

The software development and testing were primarily conducted at \textit{DFKI-RIC} in Bremen, Germany, with the final software version deployed on-site. As the robots were shipped about 2 weeks before the team traveled, a branch of the software that had been last tested on the hardware was frozen. Further developments were carried out on a separate branch, tested in simulation, and was merged into the frozen field test branch feature-by-feature after the team arrived on-site. This approach ensured that the software was stable and compatible with the hardware, minimizing potential issues during field deployment. This allowed the team to test and merge all the new features within half a day after arriving on-site. For the software architecture, we used ROS 2 Humble on Ubuntu 22.04, which provided a robust framework for function and callback timers, communication middleware, and data management tools such as logging and plotting. The software was containerized using Docker to ensure consistency across different development machines and ease of deployment. We also benchmarked the controller's real-time performance on different computing architectures (x86 and ARM) to ensure that the software could run efficiently on the available hardware and enable comparisons between different hardware setups. The software stack can be seen in \autoref{fig:sw_stack}. The results of these benchmarks and the open source software can be found in \cite{stark2025icra, quadgithub}.

A robust network setup was established using a local Wi-Fi router to facilitate communication between the robots and the ground station. This setup allowed for real-time monitoring and control of the robots during field tests. Additionally, a separate cellular connection was used for the D-GPS base station to ensure reliable GPS corrections, which are crucial for accurate localization in outdoor environments. The team also prepared for potential connectivity issues by having offline data logging capabilities on the robots. All the routers and the ground station control laptops were powered using large USB-C power banks, as there was no power available on-site. This setup ensured continuous operation during the field tests, allowing for extended periods of data collection and analysis without interruption. The USB-C power banks were within the limits of air travel, which was important for transporting them to the site.

\subsection{On Field Activities}

\begin{figure}[t]
    \centering
    \includegraphics[width=\linewidth]{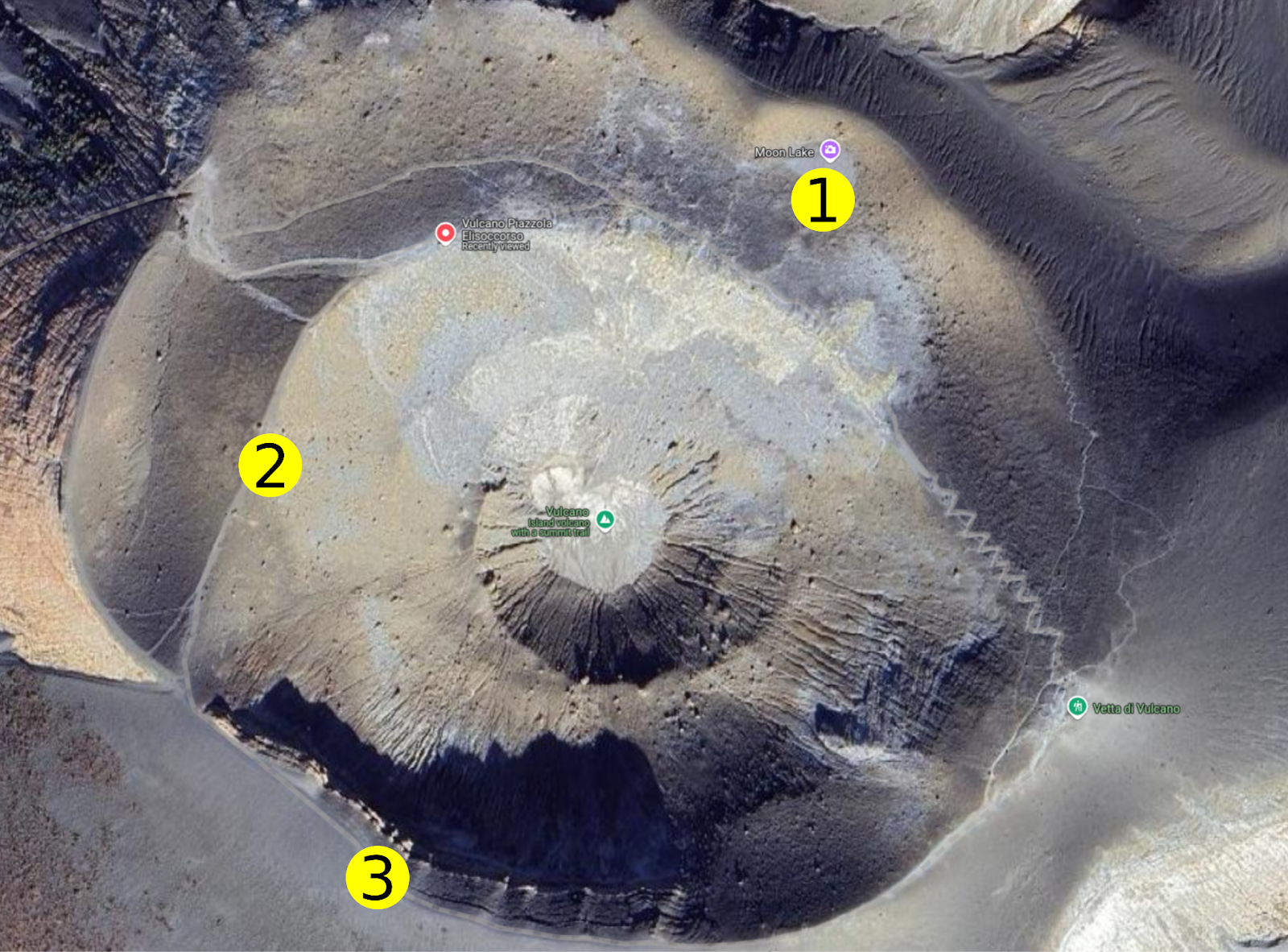}
    \caption{Aerial/Satellite view of Vulcano island with the marked test sites. 1: Moon Crater/Lunar Lake, 2: Lower Rim, 3: Upper Rim.}
    \label{fig:test_sites}
\end{figure}

\begin{figure*}[bht]
    \centering
    \includegraphics[width=\textwidth]{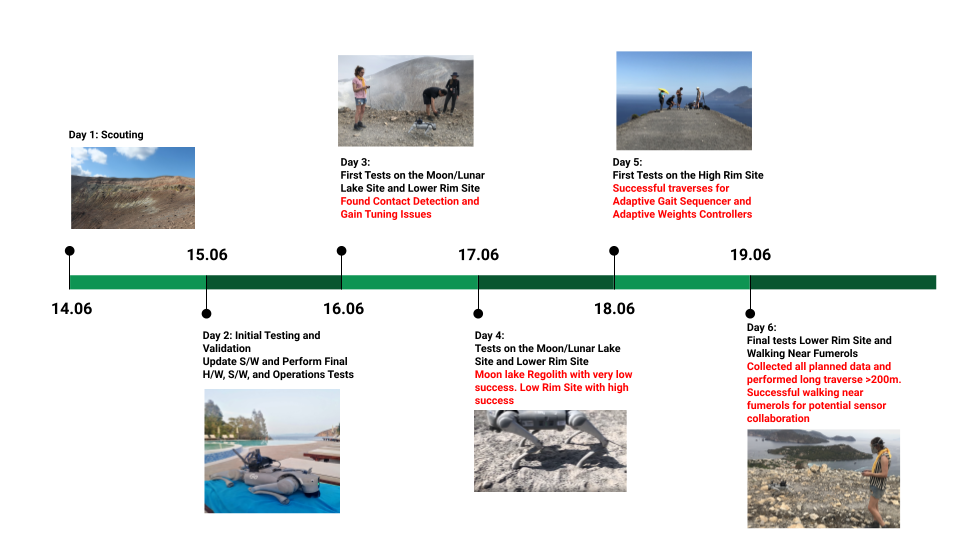}
    \caption{Timeline of Activities during the field campaign on Vulcano.}
    \label{fig:timeline}
\end{figure*}

During the field tests, the team followed a structured daily schedule to maximize productivity and ensure safety. The timeline of the activities is shown in Fig.~\ref{fig:timeline}. The team traveled to the Vulcano at 6:00 AM each day to take advantage of the cooler morning temperatures for both the hike to the top, and avoiding overheating during robot operation. The tests were carried out until 11:00 AM, after which the team took a break for lunch and rest during the hottest part of the day. This was to avoid overheating of the robot during the harsh summer sun combined with warm volcanic ground and to reduce the chance of sunburns for the team. The tests were carried out at 3 test sites: Moon Crater/Lunar Lake, Lower Rim, and Upper Rim (see Fig.~\ref{fig:test_sites}). The selection of the test sites for each day depended on the weather conditions and volcanic activity reports i.e. to avoid sites with volcanic gasses coming out from fumaroles and being blown by the wind. The first day was primarily dedicated to controller tuning rather than research data collection, underscoring the importance of allocating sufficient time for system setup and calibration. The following test days focused on data collection and validation of the control algorithms. A typical test plan is shown in Table~\ref{tab:test_plan}. The tests included a variety of locomotion tasks, such as straight-line traversal at different speeds, zigzag patterns, and long traverses to evaluate the robot's performance under different conditions.

\begin{table}[thpb]
\centering
\resizebox{\linewidth}{!}{
    \begin{tabular}{r|l|c|c|c|c|c|l}
        \textbf{id} & \textbf{Gait sequencer} & \textbf{Gait} & \textbf{Velocity} & \textbf{Test site} & \textbf{Distance} & \textbf{Fail after} & \textbf{Description} \\\hline
        1.1 &Adaptive & Trot & free & 2 & 4*10\,m & -- & \multirow{5}{*}{\shortstack[l]{Direction\\ control, \\ user speed}}\\
        1.2 &Simple   & Trot & free & 2 & 10\,m & 7\,m & \\
        1.3 &         & Trot & free & 2 & 10\,m & 1\,m & \\
        1.4 &         & Trot & free & 2 & 10\,m & 0.5\,m & \\
        1.5 &         & Trot & free & 2 & 10\,m & 7\,m & \\\hline
        2.1 &Adaptive & Trot & 0.1\,m/s & 2 & 10\,m & -- & \multirow{8}{*}{\shortstack[l]{Straight line,\\ fixed speed}}\\
        2.2 &         & Trot & 0.2\,m/s & 2 & 10\,m & --& \\
        2.3 &         & Trot & 0.3\,m/s & 2 & 10\,m & --& \\
        2.4 &         & Trot & 0.4\,m/s & 2 & 10\,m & --& \\
        2.5 &Simple   & Trot & 0.1\,m/s & 2 & 10\,m & 0.5\,m& \\
        2.6 &         & Trot & 0.2\,m/s & 2 & 10\,m & 10\,m& \\
        2.7 &         & Trot & 0.3\,m/s & 2 & 10\,m & 5\,m& \\
        2.8 &         & Trot & 0.4\,m/s & 2 & 10\,m & 2\,m& \\\hline
        3.1 &Adaptive & Trot & 0.2\,m/s + free & 1 & 10+10\,m & -- & \multirow{5}{*}{\shortstack[l]{Straight line,\\zigzag on \\way back}}\\
        3.2 &         & Walk & 0.2\,m/s + free & 1 & 10+10\,m & -- & \\
        3.3 &Simple   & Trot & 0.2\,m/s + free & 1 & 10+10\,m & 16\,m & \\
        3.4 &         & Walk & 0.2\,m/s + free & 1 & 10+10\,m & 10\,m & \\
        3.5 &         & Walk & 0.2\,m/s + free & 1 & 10+10\,m & -- & \\\hline
        4.1 &Adaptive & Trot & free & 2 & 95\,m & -- &\multirow{3}{*}{\shortstack[l]{Long\\ traverse, \\user speed}} \\
        4.2 &         & Trot & free & 2 & -- & 60\,m & \\
        4.3 &Simple   & Trot & free & 2 & -- & 17\,m & \\\hline
        5.1 &Adaptive & Trot & free & 3 & 10\,m & -- & \multirow{4}{*}{\shortstack[l]{Direction\\control,  \\user speed}}\\
        5.2 &         & Trot & free & 3 & 10\,m & 2\,m & \\
        5.3 &         & Walk & free & 3 & 10\,m & 3\,m & \\
        5.4 &Simple & Trot & free & 3 & 10\,m & 1\,m & \\
    \end{tabular}
}
\caption{Sample test plan from the field tests}
\label{tab:test_plan}
\end{table}

\begin{figure}[htpb]
    \centering
    \includegraphics[width=\linewidth]{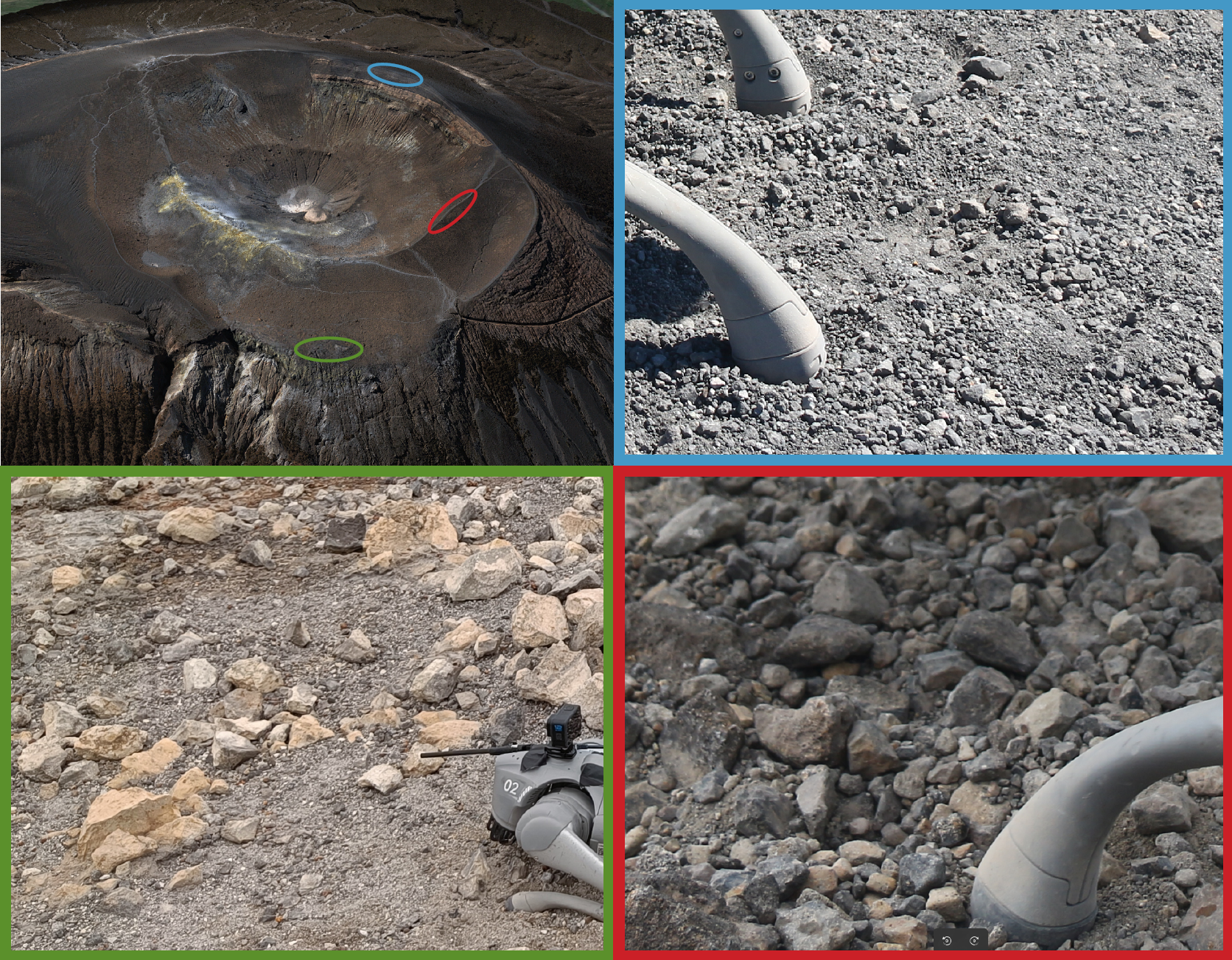}
    \caption{Top left: Aerial View of Volcano and test sites. Other images show different regoliths, terrains, and volcanic features encountered during the tests.}
    \label{fig:vulcano_collage}
\end{figure}

During the tests, the robot encountered various terrains and volcanic features, as shown in Fig.~\ref{fig:vulcano_collage}. The challenging terrain, particularly the fine regolith, posed traction difficulties for the robots; nevertheless, their performance exceeded expectations, although terrain-induced challenges persisted throughout the tests. Accessing more hazardous terrain will require gas masks in future campaigns. Over the course of the campaign, the team successfully traversed over 700 meters and collected extensive data, marking the first recorded quadrupedal walking on an active volcano. The team ensured that all safety protocols were followed, especially given the active volcanic nature of the site. This included constant monitoring of volcanic activity reports and adherence to guidelines provided by local scientists and volcanologists.

\section{Adaptive Control Architecture}
\label{sec:control_architecture}

\begin{figure}[bhtp]
    \centering
    \includegraphics[width=\linewidth]{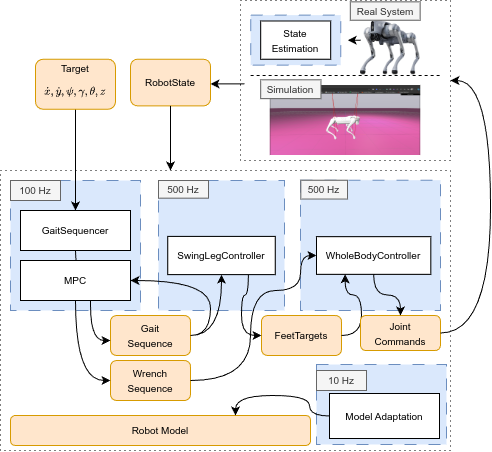}
    \caption{Control Architecture used during the field tests on Vulcano.}
    \label{fig:control_architecture}
\end{figure}

The control of the quadruped robot is based on a hierarchical architecture that combines a high-level gait planner with a low-level model predictive controller (MPC) and an instantaneous stabilization controller. This is similar to the control architecture developed for the \textit{mini-cheetah} robot \cite{kim2019highly}. The controller pipeline consists of three main components: a gait planner, a model predictive controller (MPC), and an instantaneous stabilization controller using Whole Body Control (WBC). The overall control architecture is depicted in Fig.~\ref{fig:control_architecture}. The high-level gait planner generates desired footfall patterns and body trajectories based on the selected gait and desired velocity. The single rigid body MPC optimizes the center of mass's motion by solving a constrained optimization problem that considers the centroidal dynamics and contact constraints. This outputs the contact forces for the required motion. The whole body controller maps the contact forces to the joint torques while ensuring the robot's stability and adherence to physical constraints. The control architecture is implemented in ROS 2, allowing for modularity and ease of integration with other system components. The adaptive elements of the controller enable real-time adjustments to changing terrain conditions, enhancing the robot's ability to maintain stability and performance during locomotion on uneven and unpredictable surfaces. The control software is open-sourced and can be found in \cite{quadgithub}.

The control software was made adaptive on various levels: the gait planner was bio-inspired and could adapt the gait parameters based on the speed of the robot; the MPC ran along with an online model adaptation to allow for changing of the system model based on attached payloads; and the state estimation adapted the covariances used in the Kalman filter online based on the performance of the system. An in-depth description of the adaptive control methods used can be found in \cite{rohit2025astra}.

\section{Lessons Learnt}
\label{sec:lessons_learnt}

The field campaign on Vulcano yielded several important insights for future planetary robotics deployments. The campaign was highly successful, with over 700 meters traversed and extensive data collected, marking the first recorded quadrupedal walking on an active volcano. The challenging terrain, particularly the fine regolith, posed traction difficulties for the robots; nevertheless, their performance exceeded expectations, although terrain-induced challenges persisted throughout the tests. Adaptive control algorithms demonstrated strong capabilities, but further improvements are necessary to achieve greater robustness in unpredictable environments. State estimation was complicated by the absence of a compass in the Go2 robot's IMU, which prevented alignment with D-GPS measurements and limited benchmarking to absolute distance only. Reliable communication was ensured by a custom network setup using a mobile Wi-Fi router, while a separate cellular connection for the D-GPS base station enabled effective GPS corrections; having spare antennas proved critical after one was damaged during experiments. Large USB-C power banks were indispensable for powering ground station laptops, supporting continuous operation during the campaign. The first field day was primarily dedicated to controller tuning rather than research data collection, underscoring the importance of allocating sufficient time for system setup and calibration.

While the sim-to-real gap is well known in robotics, this campaign highlighted that a significant lab-to-field gap also exists: algorithms and hardware that perform reliably in controlled laboratory settings may encounter unforeseen challenges in real-world field environments, requiring additional adaptation and validation.

Accessing more hazardous terrain will require gas masks in future campaigns. Immediate validation of field test data was facilitated by basic rosbag processing scripts, allowing for same-day analysis and rapid iteration. Clear separation of "office" and "non-office" hours helped maintain team focus and efficiency. Having a dedicated person for documentation and photos, 3D printed spare parts for the robot, two robots with multiple extra batteries, and prior field experience all contributed to the success of the campaign and will make future deployments easier through improved logistics and planning.

\section{Conclusion}
\label{sec:conclusion}

This field campaign on Vulcano demonstrated the feasibility and advantages of deploying quadrupedal robots for planetary exploration in challenging analog environments. By validating adaptive control algorithms and increasing the technology readiness level (TRL) of the developed GNC sub-system, we bridged not only the sim-to-real gap but also highlighted the significant lab-to-field gap that must be addressed for future missions. The robots successfully traversed over 700 meters on analog regolith, providing valuable insights into locomotion performance and system robustness, though further extensive field trials and low-gravity testing are needed to reach higher TRLs. The campaign fostered new collaborations with geologists and planetary scientists, and provided essential experience in planning and executing field deployments for legged robots. These results support the integration of legged mobility into future lunar and Martian mission architectures and lay the groundwork for continued interdisciplinary research and technology maturation.

\section*{Acknowledgments}

This work partially funded by the projects: AAPLE (grant number 50WK2275) funded by the German Federal Ministry for Economic Affairs and Climate Action (BMWK), M-Rock (grant number 01IW21002) funded by the German Federal Ministry for Economic Affairs and Climate Action (BMWK) and the Ministry of Education and Research (BMBF), and ActGPT (grant number 01IW25002) funded by the Federal Ministry of Research, Technology and Space (BMFTR) and is supported with funds from the federal state of Bremen for setting up the Underactuated Robotics Lab (265/004-08-02-02-30365/2024-102966/2024-740847/2024). We would also like to acknowledge the support of \textit{mybotshop.de} for the transportation of the robot to the field site.

\bibliography{references}

\end{document}